\documentclass[letterpaper,conference]{IEEEtran}

\usepackage[utf8]{inputenc}
\usepackage[T1]{fontenc}
\usepackage{cite}
\usepackage{amsmath,amssymb,amsfonts}
\usepackage{graphicx}
\usepackage{booktabs}
\usepackage{multirow}
\usepackage{xcolor}
\usepackage{url}
\usepackage{siunitx}
\usepackage{subcaption}
\usepackage{enumitem}
\usepackage{geometry}
\title{\vspace{.4cm}
A Hierarchical End-of-Turn Model with Primary Speaker Segmentation\\for Real-Time Conversational AI}
\author{%
\IEEEauthorblockN{Karim Helwani, Hoang Do, James Luan, and Sriram Srinivasan\\ Meta, Menlo Park, CA, USA\\
Email: \texttt{\{karimhelwani, hdo, jamesluan, sriramsri\}@meta.com}
}
}

\begin{document}

\maketitle

\begin{abstract}
We present a real-time front-end for voice-based conversational AI to
enable natural turn-taking in two-speaker scenarios by combining primary
speaker segmentation with hierarchical End-of-Turn (EOT) detection. To operate robustly
in multi-speaker environments, the system continuously identifies and
tracks the primary user, ensuring that downstream EOT decisions are not
confounded by background conversations. The tracked activity segments are fed to a hierarchical, causal EOT model that predicts the immediate conversational state by independently analyzing per-speaker speech features from both the primary speaker and the bot. Simultaneously, the model anticipates near-future states ($t{+}10/20/30$\,ms) through probabilistic predictions that are aware of the conversation partner's speech.
Task-specific knowledge distillation compresses wav2vec~2.0
representations (768\,D) into a compact MFCC-based student
(32\,D) for efficient deployment. The system achieves 82\% multi-class
frame-level F1 and 70.6\% F1 on Backchannel detection, with 69.3\% F1 on a
binary Final vs.\ Others task. On an end-to-end turn-detection
benchmark, our model reaches 87.7\% recall vs.\ 58.9\% for Smart
Turn~v3 while keeping a median detection latency of 36\,ms versus
800--1300\,ms. Despite using only 1.14\,M parameters, the proposed model
matches or exceeds transformer-based baselines while substantially
reducing latency and memory footprint, making it suitable for edge
deployment\footnote{Paper accepted for presentation at the IEEE Conference on Artificial Intelligence 2026.}.\\\\
\end{abstract}

\section{Introduction}
\label{sec:intro}
The growing popularity of voice assistants has accelerated the adoption of voice as an intuitive front-end for LLM-powered conversational AI systems. This shift presents the significant challenge of supporting natural turn-taking in conversations, requiring systems to be responsive to user interruptions while remaining robust against background speakers and environmental noise, and to accurately interpret backchannels (such as "mm-hmm" or "uh-huh") and other natural user reactions that do not indicate the end of a conversational turn, all without depending on explicit keywords that can disrupt the conversational flow. Speaker diarization, determining ``who spoke when'', has advanced
rapidly with deep learning~\cite{park22_diarization_review}. While hybrid
pipelines that segment audio, extract speaker embeddings, and cluster them,
using encoders such as x-vectors and ECAPA-TDNN with modern clustering
like VBx~\cite{landini22_vbx,desplanques20_ecapa} are effective in single-speaker or low-overlap conditions, performance in conditions with substantial
overlap remains challenging. End-to-end neural diarization (EEND) predicts multi-speaker activity directly and handles overlap via permutation-invariant training~\cite{fujita19_eend}. Target-speaker voice activity detection (TS-VAD) further improves robustness in practical scenarios by conditioning on speaker-specific embeddings~\cite{medennikov20_tsvad}. Recent work has also emphasized augmentation and consistency learning for robust audio classification~\cite{iqbal21_consistency}, as well as latent disentanglement for structured acoustic scenes~\cite{helwani24_blockwise}.

\noindent \textbf{Contributions:} Motivated by these recent developments, we present an approach that integrates a streaming primary speaker segmentation module with a hierarchical end-of-turn (EOT) detector. Compared to prior diarization, endpointing, and turn-taking work, this paper makes the following contributions: 
\begin{enumerate}[label=(\roman*)]
\item \emph{streaming primary-speaker segmentation} that extracts disentangled multi-speaker embeddings in real-time and performs online clustering to reliably track the active user even in the presence of background speech, without requiring prior enrollment, and ensuring that only the primary user's activity is forwarded to the chatbot.
\item \emph{Hierarchical EOT} architecture with integrated backchannel detection. The EOT model operates across multiple horizons: it enforces strict causality at horizon~0 by relying exclusively on self-features, while also making predictions that are aware of the conversation partner (i.e., the other speaker) for near-future states ($t{+}10/20/30$\,ms). This architecture is explicitly designed to prevent the use of the partner’s speech onset as a shortcut for predicting turn ends in conversational datasets.
\item ~\emph{anti-shortcut data augmentation} for turn-taking, including label time-shift jitter and random channel dropout, which prevent the model from trivially memorizing conversation dynamics such as ``A ends exactly when B starts''.
\item \emph{task-specific distillation} from wav2vec~2.0's 768-D representation to a compact MFCC-based student, enabling a 32-D per-speaker representation with only 1.14\,M parameters and $\approx$111\,MMAC/s.
\item experiments on a benchmark dataset showing our system achieves higher turn-end recall with substantially lower latency and smaller model size than transformer baselines.
\end{enumerate}
\section{Related Work}
\label{sec:related}

\subsection{Endpointing and Turn-taking}

Endpointing indicates the end of a user prompt that the chatbot can process once a user has finished speaking,
traditionally using VAD-style trailing-silence rules or learned
\emph{neural endpointers}. Acoustic GLDNN/CLDNN endpointers for
streaming ASR~\cite{chang17_gridlstm_epd} and end-to-end endpoint
detectors that integrate acoustic and language modeling
knowledge~\cite{hwang20_e2e_epd} have been widely adopted. More recent
work folds endpointing into end-to-end recognizers, e.g., joint
endpointing and decoding or EOS tokens in RNN-T models, to improve the
latency--WER trade-off~\cite{chang19_joint_epd,lu22_surt_epd}. These
approaches optimize utterance termination but do not explicitly model
the mutual interaction cues that govern turn-taking. Predictive turn-taking models instead forecast near-future activity of
interlocutors. Voice Activity Projection
(VAP)~\cite{ekstedt22_vap} learns future voice-activity windows and
captures turn-shift/backchannel patterns; subsequent work extends VAP to
multilingual dialogue and investigates prosodic
cues~\cite{inoue24_multilingual_vap}. In our experiments we benchmark
against Smart~Turn~v3, an open-source ``semantic VAD'' that predicts
turn completion from raw audio~\cite{smartturnv2}. Unlike VAP or our
hierarchical approach, Smart~Turn does not explicitly model bi-speaker
future activity or multi-horizon causal structure, making it a strong baseline but not a predictive turn-taking model per se.

\subsection{Diarization and Primary Speaker Tracking}

Diarization pipelines commonly combine frame-level segmentation,
embedding extraction (x-vector, ECAPA), and clustering
(PLDA/AHC, VBx)~\cite{park22_diarization_review,landini22_vbx,desplanques20_ecapa}.
EEND~\cite{fujita19_eend} predicts overlapped activities jointly using PIT
loss, while TS-VAD~\cite{medennikov20_tsvad} conditions on target embeddings
to improve detection. Our focus is an online, streaming variant tailored to conversational AI: the goal is not complete diarization, but reliable filtering of non-primary talkers so the downstream EOT model operates on the intended speaker only. We adopt causal convolutions for streaming, multi-condition augmentation and consistency learning~\cite{iqbal21_consistency,hendrycks20_augmix}, and a lightweight disentanglement-inspired peeling mechanism to obtain multiple concurrent speaker embeddings~\cite{helwani24_blockwise}.

\subsection{Relation to VAP and Smart Turn}\label{ssec:rel_vap}

Voice Activity Projection (VAP)~\cite{ekstedt22_vap} predicts future joint voice-activity trajectories for both speakers and can be used to derive turn-shift and backchannel events. In its original form, VAP processes bi-speaker audio jointly through a shared encoder, so the model leverages the instantaneous activity of the conversation partner as a strong cue for turn-ends; this is natural for offline analysis, but in a real-time system where turn-end decisions \emph{trigger} the partner's response, it risks learning a non-causal shortcut (``A has finished because B has started''). Our architecture instead enforces that horizon-0 decisions for each speaker use only that speaker's features, with partner information entering only in later horizons and through explicit anti-shortcut augmentation. Smart~Turn~v3~\cite{smartturnv2} is a compact ``semantic VAD'' that operates directly on the waveform and predicts turn completion at the utterance level. It serves as a strong engineering baseline in our experiments, but unlike our model it does not (i) track a primary speaker in multi-talker mixtures, (ii) model multi-horizon turn states (Initial/Speech/Interim/Final/Backchannel), or (iii) provide strictly causal frame-level predictions suitable for sub-50\,ms turn-taking.

\section{Primary Speaker Segmentation}
\label{sec:primary}

\subsection{Motivation}

In multi-speaker environments, while a VAD is effective in rejecting ambient background noise, it cannot separate the engaged user from other background talkers. Without explicit primary user filtering, EOT triggers on irrelevant speech, causing spurious handovers. We therefore segment and track the primary user online and pass only their activity to the EOT detector.

\subsection{Feature Encoder (Backbone)}

Input audio is converted to 32-D MFCCs with a 20\,ms window,
10\,ms hop, and causal framing (100\,Hz). A linear projection maps
frames to 128\,D, followed by $L$ Multi-Scale Causal Convolution (MSCC)
blocks. Each block applies 1-D convolutions at multiple dilations
($d \in \{1,4\}$) with left padding $(k{-}1)d$ for causality and
fuses branches via a $1{\times}1$ convolution and dropout: 
$ h^{(\ell)}_t = \mathrm{ReLU}\!\left(\mathrm{BN}\big(\mathrm{Conv1D}_\ell(x)\big)\right).$
Stacking such blocks yields a lightweight causal encoder reminiscent of
ECAPA-TDNN~\cite{desplanques20_ecapa} but optimized for streaming
speaker tracking.

\subsection{Peeling-based Multi-speaker Embedding Extraction}

To handle overlapping speakers, we iteratively extract $K$ disentangled embedding
sequences $\{z_1,\dots,z_K\}$ via a ``peeling'' mechanism inspired by latent
block-wise disentanglement~\cite{helwani24_blockwise}. Let $r_0$ denote
the encoder output after passing a fully connected layer serving as a residual projection layer. For iteration $i \in \{1,\dots,K\}$:
\begin{align}
h_i &= \tanh(W_e r_{i-1}), \quad
g_i = \sigma(W_g r_{i-1}), \\
z_i &= \mathrm{SoftQuant}\big(128\,(g_i \odot h_i)\big)/128, \\
s_i &= \sigma(W_s r_{i-1}), \quad
r_i = r_{i-1} - z_i,
\end{align}
where $z_i$ is the $i$-th speaker embedding sequence, $s_i$ is a stop
signal that informs the number of speakers in a given frame, $r_i$ the residual signal, $W_e, W_g, W_s$ refer to the weights of the embedding, gate, and stop signal heads respectively. The residual update encourages successive peels to model distinct sources. After $K$ iterations we obtain
$Z = \{z_1,\dots,z_K\}$.
\begin{figure*}
    \centering
    \includegraphics[width=0.5\linewidth]{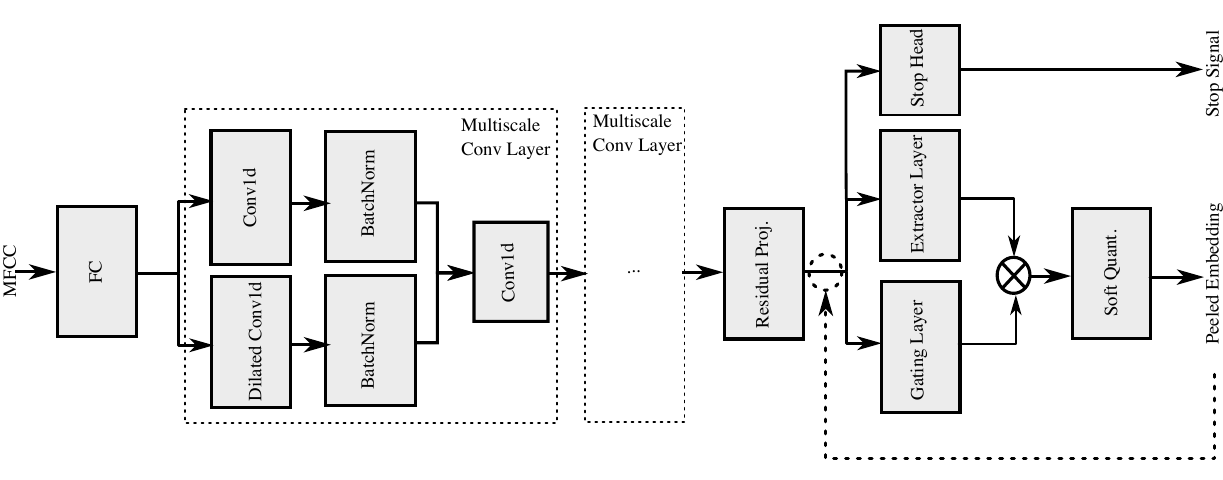}
    \caption{Detailed overview of the primary speaker segmentation unit. The embeddings are extracted recursively, the gating layer decides how much of the residual signal $r_i$ to be used for a speaker' embedding, the number of speakers in a frame is informed through the stop signal $s_i$. The embeddings are quantized using a soft quantizer into 128 discrete levels.}
    \label{fig:overview}
\end{figure*}
\subsection{Online Primary Speaker Clustering}

The centroid for the primary speaker is updated online using a momentum-based moving average whenever the cosine similarity between the incoming embedding $z_t$ and the current centroid $c_{\mathrm{prim}}$ exceeds a threshold:
\begin{align}
\rho_t &= \cos(z_t, c_{\mathrm{prim}}), \\
\text{if } \rho_t &> \theta:\quad
c_{\mathrm{prim}} \leftarrow
\alpha c_{\mathrm{prim}} + (1-\alpha) z_t,
\end{align}
with threshold $\theta{=}0.7$ and momentum $\alpha{=}0.9$ determined empirically. The centroid
initializes on the first active frames observed from the engaged user
(e.g., after first talker speech activity following agent prompt) and adapts smoothly, maintaining
robustness to brief background intrusions. Only frames whose
peel-embeddings are closest to $c_{\mathrm{prim}}$ (and pass VAD) are
forwarded to the EOT module.

\subsection{Training Objectives and Augmentation}

We combine permutation-invariant supervision for overlapped peels,
margin-based discrimination, and consistency regularization:

\begin{itemize}
  \item \textbf{PIT binary cross-entropy} over peels and reference masks
        (EEND-style) to ensure permutation robustness~\cite{fujita19_eend}.
  \item \textbf{CosFace margin loss} to separate speakers
        angularly~\cite{wang18_cosface}.
  \item \textbf{Supervised contrastive loss} to pull same-speaker frames
        together and push different speakers apart~\cite{khosla20_supcon}.
  \item \textbf{VAD-masked triplet consistency} between clean and
        mixture embeddings, aligning mixture peels with clean references.
  \item \textbf{Peel diversity loss} penalizing high cosine similarity
        between different peels, reducing redundancy.
  \item \textbf{AugMix-style prediction consistency} across augmentations
        using Jensen--Shannon divergence~\cite{hendrycks20_augmix,iqbal21_consistency}.
\end{itemize}

We employ multi-condition augmentation: additive noise/music
(SNR 10--15\,dB), RIR reverberation with a reverberation time ($T_{60}$ 100--800\,ms), PSTN downsample/$\mu$-law
upsampling, Wiener AEC artifacts, and overlap simulation by mixing two
speakers with random ratio $\alpha \in [0.3,0.7]$. A Silero VAD masks
non-speech frames during training~\cite{silero_vad}.

\subsection{Performance Evaluation and Model Complexity}

\begin{figure}[t]
  \centering
  \includegraphics[width=.7\linewidth]{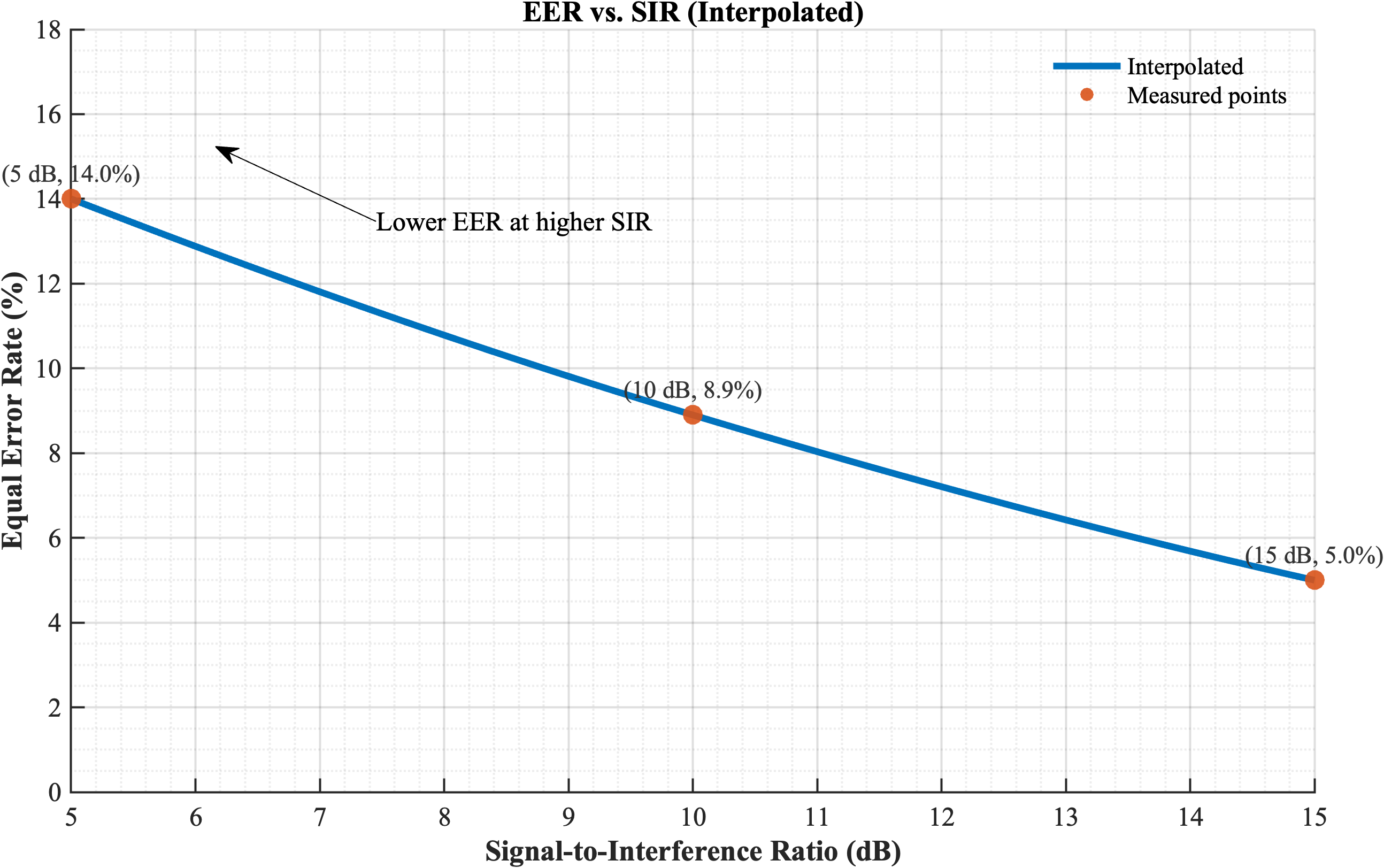}
  \caption{Equal error rate (EER) vs.\ signal-to-interference ratio
  (SIR) for primary speaker segmentation. Average EER is $\approx$10\%,
  improving as SIR increases.}
  \label{fig:eersir}
\end{figure}

On mixtures with varying signal-to-inference-ratios (SIRs), the primary segmentation achieves an
average equal error rate (EER) of $\approx$10\%, improving to $<8$\% for
SIR${>}10$\,dB and degrading gracefully at low SIRs (Fig.~\ref{fig:eersir}). This
effectively filters background speech prior to EOT, reducing false
triggers from non-primary talkers. The online clustering, and feature and embedding extractions have a median end-to-end detection latency from the speech onset of 50 ms. The primary segmentation model is light weight requiring $\approx 137$\,k MACs.

\vspace{-0.5cm}

\section{Hierarchical End-of-Turn Model}
\label{sec:model}

We perform per-frame (100\,Hz) EOT state classification for two
speakers. Each 10 ms frame contains 46 features: 40 MFCCs (20 per speaker),
4 pitch features (2 per speaker), and 2 voice-activity indicators
(1 per speaker). The model predicts five classes per speaker:
\emph{Initial}; representing the silent frames before a turn starts, \emph{Speech} for talker speech activity frames, \emph{Interim}; for pauses the speaker makes within a turn, \emph{Final}; representing the end of a turn, and
\emph{Backchannel}; representing non-interrupting conversational cues. We assess performance using both frame-level classification metrics (including multi-class F1 and accuracy, as well as binary F1 for distinguishing \emph{Final} versus Others) and turn-level detection metrics, which are computed by applying a threshold to the aggregated \emph{Final} probability across conversational segments. In the following sub-sections, we describe the labeled dataset we use, the teacher-student model architecture, and discuss the computational complexity of the model. 

\subsection{Datasets and Labels}

We use $\sim$350 hours of human annotated proprietary dataset of segment/(turn)-level diarized English conversational speech for training, which we convert to frame-level (100\,Hz) labels. Each frame is
assigned one of five classes per speaker: \emph{Initial},
\emph{Speech}, \emph{Interim}, \emph{Final}, or \emph{Backchannel}. Speech frames inside an annotated segment are labeled as \emph{Speech};
pauses within a segment (turn) (audio VAD is 0) are
labeled as \emph{Interim}. Backchannels are detected via lexical and
durational heuristics applied to the transcription (e.g.\ short
utterances such as ``mm-hm'', ``yeah'', ``right''), and the corresponding
frames are labeled as \emph{Backchannel}. To obtain frame-level \emph{Final} labels, we work at turn level. A
\emph{complete turn} is annotated by a human labeler. For each turn, let $t_{\text{end}}$ be the frame
index of the last speech frame of that turn. We then assign the label
\emph{Final} to a fixed window of 10 frames (100\,ms at 100\,Hz)
starting at $t_{\text{end}}$, i.e.\ frames
$[t_{\text{end}}, t_{\text{end}}+9]$ for that speaker. All other frames
remain in their respective classes (\emph{Initial}, \emph{Speech},
\emph{Interim}, \emph{Backchannel}). This provides a frame-level
approximation of the turn-end region suitable for multi-class training. The dataset contains both conversations and single-speaker segments. However, single-speaker segments used primarily to pre-train the
distillation student, encouraging self-conditioned Final predictions.

\subsection{Model Overview}

Stereo audio at 16\,kHz is converted to frame-level features at 100\,Hz. For each speaker we compute 20 MFCCs, pitch (F0 and voicing
probability), and 1-D voice-activity indicators. During training we compute wav2vec~2.0
features~\cite{baevski20_wav2vec2} for each speaker and pass them
through a small compression (teacher) network (768$\rightarrow$32\,D) using an MLP with residual skip (768$\rightarrow$32) and layer normalization. Compression is applied per speaker. A shared student network maps each speaker's 20-D MFCCs to a 32 D task-specific compressed representation using multiscale causal convolutions and residual connections. The student's 32\,D MFCC features are matched to the compressed teacher
via a loss combining L1, L2, and cosine similarity with weights
(0.5, 0.3, 0.2). Teacher targets are detached to prevent gradients
flowing into the compression module from the distillation term. To mitigate train--test mismatch, we gradually mix teacher and student
compressed features via a cosine decay schedule, transitioning from
100\% teacher to pure student over a warm-up window. Subsequent EOT heads consume
the mixed features during training; inference uses only the student. For each speaker, the base features (23\,D: MFCC 20 + pitch 1 + voicing 1 + VAD 1) are
concatenated with the 32\,D compressed features (55\,D total) and mapped
via a 55$\rightarrow$256$\rightarrow$128 MLP with ReLU and dropout. The 55\,D feature vector is passed to the EOT head, which is a 2-layer LSTM (hidden 128, dropout 0.1) that models temporal dynamics with
state caching for streaming. The EOT head consumes either teacher (training) or student (inference) features. Horizon-0 heads predict immediate states using only self-features, taking the speaker's LSTM output and predicting 5-class
logits (causal) corresponding to the five classes per speaker:
\emph{Initial}, \emph{Speech}, \emph{Interim}, \emph{Final}, and
\emph{Backchannel}. Future heads ($h1-h3$) incorporate adapted partner features and self horizon-0 logits, enabling anticipation while constraining cross-speaker influence. They receive a concatenation of self LSTM
output (128\,D), adapted partner features (32$\rightarrow$8\,D, dropout
0.2), and self horizon-0 logits (5\,D). A 141$\rightarrow$15 linear
layer outputs logits for horizons $h1-h3$ (3$\times$5), reshaped to
$[T,3,5]$. Per speaker, horizons $[h0,h1,h2,h3]$ are concatenated into
$[T,4,5]$ and truncated to valid positions to respect the prediction
lookahead. Outputs are stacked across speakers to form an
$[2,T,4,5]$ tensor. For training we use a Toeplitz-style loss that
unfolds the per-horizon predictions into overlapping windows and
computes cross-entropy with shifted labels, so that each frame predicts
its own state and three future frames.
The overall system overview is depicted in Fig.~\ref{fig:overview}.
\begin{figure*}
    \centering
    \includegraphics[width=0.50\linewidth]{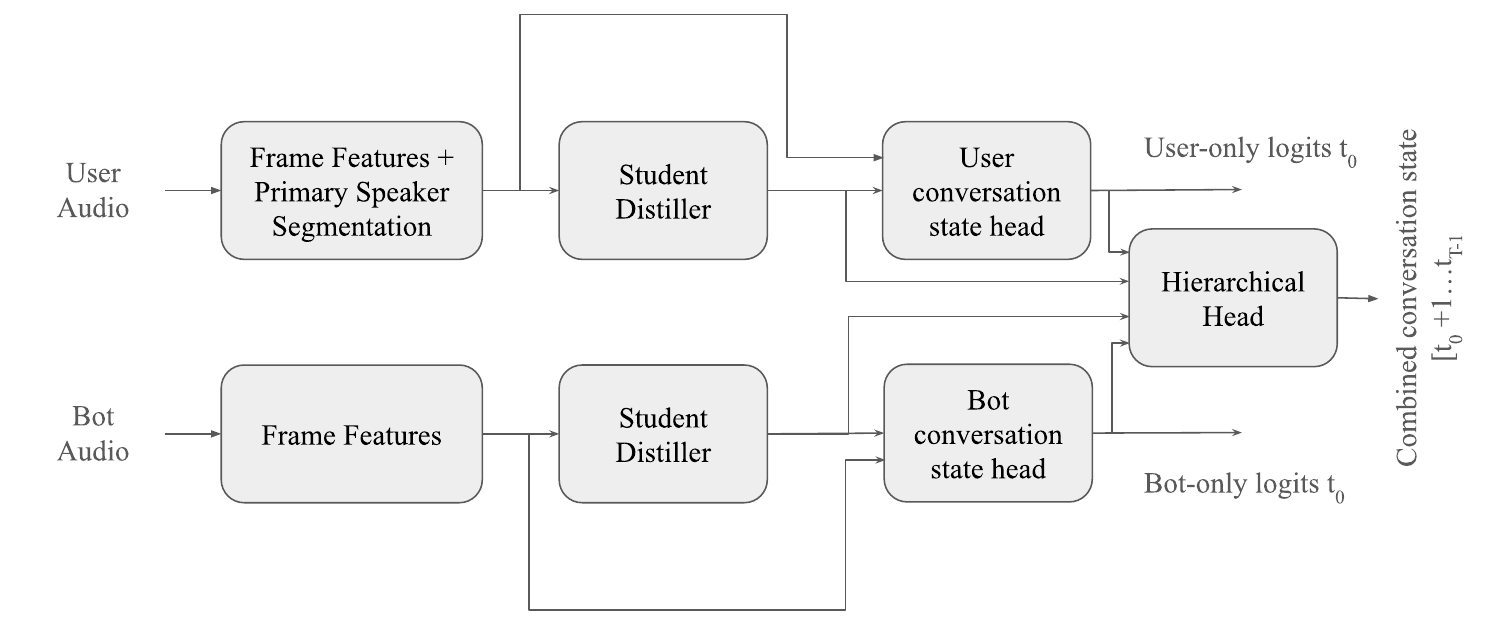}
    \caption{System Overview.}
    \label{fig:overview}
\end{figure*}
\begin{table*}[!h]
\centering
\caption{Comparison between hierarchical EOT and Smart~Turn~v3 on
357 samples. Detection latency is measured end-to-end.}
\label{tab:smartturn}
\begin{tabular}{lcccccc}
\toprule
Model & Params & Latency & Prec. & Rec. & F1 & Acc. \\
\midrule
Hierarchical EOT & 1.14M &
36\,ms & 57.2\% & 87.7\% & 69.3\% & 62.4\% \\
Smart Turn~v3~\cite{smartturnv2} & 8--12M &
900\,ms & 68.4\% & 58.9\% & 63.3\% & 69.7\% \\
\midrule
Advantage & $\sim$8--10$\times$ &
$\sim$25$\times$ & -- & +28.8\,pp & -- & -- \\
\bottomrule
\end{tabular}
\end{table*}

\subsection{Computational Complexity}
\label{sec:complexity}

At 100\,Hz, feature extraction requires roughly 720\,k MACs per frame
(dominated by FFT and mel filtering). The neural forward pass adds
$\approx$392.6\,k MACs per frame across both speakers, totaling
$\approx$1.11\,M MACs/frame ($\approx$111\,MMIPS). Weight memory is
$\approx$1.14\,M parameters ($\approx$4.56\,MB, fp32), and typical
streaming activations are $<200$\,kB. On modern CPUs, per-frame latency
is 1.5--2.0\,ms, leaving ample headroom under a 10\,ms budget.

\section{Experiments}
\label{sec:experiments}

\subsection{Training Setup}

We train on a ~350-hour proprietary dataset with frame-level labels, dropout 0.1 in standard layers and 0.2 in the
partner adapter, and cepstral mean normalization for MFCCs. Pitch
features are log-normalized. The distillation weight is 0.15 relative to
the EOT loss. We train using AdamW with warmup, gradient clipping, and
a cosine learning-rate schedule. The classification loss at each time step is cross-entropy loss with horizon weighting that prioritizes immediate predictions ($t_0=\times 1$, $t_0+10\textrm{\,ms}=\times 0.5$, $t_0+20\textrm{\,ms}=\times 0.25$,  $t_0+30\textrm{\,ms}=\times 0.1$), ground truth labels are converted to sliding windows to match each horizon and class weights are calculated to address imbalance between frequent (speech) and rare (Final, Backchannel) classes. To reduce the risk of non-causal shortcuts, we apply two
turn-taking-specific augmentations:

\textbf{Label time-shift jitter.} For a random subset of training
segments we jitter the annotated EOT boundaries by a small random offset
within a 0--1\,s window. This prevents the model from relying on
a trivial deterministic pattern such as ``speaker A always ends
exactly at the onset of speaker B'', and encourages it to use
speaker-internal cues like prosody, silence duration, and hesitation
patterns.

\textbf{Channel dropout.} Independently for each speaker channel, with
small probability we zero out the entire feature stream for that
speaker in a training segment. This forces the model to remain robust
when one channel is missing or corrupted and, crucially, discourages it
from depending exclusively on the partner channel to predict turn ends. These augmentations are applied on top of the signal-level
multi-condition augmentation used for the primary speaker segmentation
module (noise, reverberation, telephony, overlap simulation).
\subsection{Results on EOT Detection}\label{sec:protocol}
We report frame-level multi-class accuracy and F1, with emphasis on
Final detection, and include a binary Final vs.\ non-Final metric to
compare to endpoint-detection baselines. The model prediction we use for evaluation are based on majority voting of the different available horizons at a given frame. For turn-level detection, we define the detection latency for a
given turn as the time difference between the true turn-end time
$t_{\text{end}}$ (obtained from the frame-level labels) and the first
time $t_{\text{det}} \ge t_{\text{end}}$ at which the model's Final
score crosses a fixed decision threshold. Latency is thus
$t_{\text{det}} - t_{\text{end}}$, and we report the median over all
turn ends. Table~\ref{tab:smartturn} summarizes turn-level detection performance on
357 conversation samples. Our hierarchical EOT model achieves 87.7\%
recall vs.\ 58.9\% for Smart~Turn (+28.8\,pp). High recall is desirable
in real-time interaction, where missing a turn end is more detrimental
than occasional false positives; Smart~Turn is more precision-oriented. Overall multi-class average F1 is 82\%. Backchannel detection achieves
70.6\% F1, reducing false turn-switches when a decision policy suppresses
switches under high Backchannel probability. On the binary task,
Final vs.\ Others, the proposed model reaches 69.3\% F1, comparable to
transformer-based Smart~Turn (63.3\%) while operating at substantially
lower latency and parameter count. Our model provides frame-level predictions with median delay
$\approx$36\,ms on samples with clear endpoints. Smart~Turn~v3 operates on fixed-length waveform segments and produces a
single turn-completion score per segment, rather than frame-level
predictions. To evaluate it fairly on our test set, we apply it to
overlapping windows of 8\,s duration with a hop of 10 frames (100\,ms). Each window's score is assigned to the timestamp of its \emph{end}
(i.e., 8\,s after the window start). We then align these scores to our
100\,Hz frame grid by treating the score from each 8\,s window as the
Final score at its end time. For a given turn with true end time
$t_{\text{end}}$, the detection time $t_{\text{det}}$ for Smart~Turn is
defined as the end time of the first window whose end satisfies
$t_{\text{det}} \ge t_{\text{end}}$ and whose score exceeds the Final
threshold. Detection latency for Smart~Turn is then computed as
$t_{\text{det}} - t_{\text{end}}$ according to the generic definition
in Section~\ref{sec:protocol}, allowing a direct comparison with our
frame-based hierarchical EOT model. The observed median detection latency for Smart~Turn was 900\,ms. This substantial
reduction enables responsive turn-taking. Despite its smaller size
(1.14\,M vs.\ 8--12\,M parameters) and streaming operation, our model
maintains comparable overall F1 (69.3\% vs.\ 63.3\%) while delivering
lower latency and memory footprints suitable for edge deployment. While Smart~Turn attains higher overall accuracy (69.7\% vs.\ 62.4\%), it does so with markedly lower recall and much higher latency; our model is deliberately recall-oriented and low-latency to better support responsive conversational agents.

\section{Conclusion}
\label{sec:conclusion}

We introduced a hierarchical, causal, and compact EOT model with explicit
Backchannel modeling and multi-horizon prediction, coupled with a
streaming primary speaker segmentation module. The segmentation module
disentangles and clusters speaker embeddings online to forward only the
primary user's activity. Task-specific distillation yields a 32-D
MFCC-based student that matches compressed wav2vec~2.0 features while
remaining lightweight. The system reaches strong accuracy at
$\approx$111\,MMIPS, with $\approx$1.14\,M parameters and $<$200\,kB
streaming activations, enabling real-time edge deployment. Future work includes multi-lingual extensions, integration with
text-based signals from ASR and LLMs, and tighter coupling between EOT
and endpoint-only detectors for hybrid turn-taking strategies.

\section*{Acknowledgment}

This work made use of generative AI tools (LLAMA 4 Maverick \cite{llama4} 
to assist with language editing and phrasing; all technical content, experiments,
and conclusions are the authors' own.

\bibliographystyle{IEEEtran}
\bibliography{refs}

\end{document}